    \documentclass[10pt,twocolumn,letterpaper]{article}
    
    \usepackage[pagenumbers]{cvpr} 
    
    \usepackage{graphicx}
    \usepackage{multicol}
    
    \usepackage{amsmath}
    \usepackage{amssymb}
    \usepackage{booktabs}
    \usepackage[utf8]{inputenc} 
    \usepackage[T1]{fontenc}    
    \usepackage{url}            
    \usepackage{booktabs}       
    \usepackage{amsfonts}       
    \usepackage{nicefrac}       
    \usepackage{microtype}      
    \usepackage{lipsum}
    \usepackage{fancyhdr}       
    \graphicspath{{media/}}     
    \usepackage{algorithm}
    \usepackage{algorithmic}
    \usepackage{cite}
    \usepackage{appendix}
    \usepackage{listings}
    \usepackage{xcolor}
    \usepackage{multirow}
    \usepackage{array}
    \usepackage{diagbox} 
    \usepackage{wrapfig}
    
    \usepackage[pagebackref,breaklinks,colorlinks]{hyperref}

    \usepackage[capitalize]{cleveref}
    \crefname{section}{Sec.}{Secs.}
    \Crefname{section}{Section}{Sections}
    \Crefname{table}{Table}{Tables}
    \crefname{table}{Tab.}{Tabs.}

    \def\cvprPaperID{20241130} 
    \def\confName{JFR}
    \def\confYear{2024}

\begin{document}
    
    \title{Grasp What You Want: \\
    Embodied Dexterous Grasping System Driven by Your Voice}
    
    \author{
        Junliang Li$^{1,2}$\thanks{Equal First Author.} \quad Kai Ye$^{1,2}$\footnotemark[1] \quad Haolan Kang$^{1,3}$\thanks{Equal Second Author.} \quad Mingxuan Liang$^{1}$\footnotemark[2] \\
        Yuhang Wu$^{1,3}$\footnotemark[2] \quad Zhenhua Liu$^{1}$ \quad Huiping Zhuang$^{3}$ \quad Rui Huang$^{1,2}$\thanks{Corresponding authors.} \quad Yongquan Chen$^{1,2}$\footnotemark[3] \\
        \\
        $^{1}$Shenzhen Institute of Artificial Intelligence and Robotics for Society, China \\
        $^{2}$The Chinese University of Hong Kong, Shenzhen, China \\
        $^{3}$South China University of Technology, China \\
    }

    \maketitle
    
    \begin{abstract}
    In recent years, as robotics has advanced, human-robot collaboration has gained increasing importance. However, current robots struggle to fully and accurately interpret human intentions from voice commands alone. Traditional gripper and suction systems often fail to interact naturally with humans, lack advanced manipulation capabilities, and are not adaptable to diverse tasks, especially in unstructured environments. This paper introduces the Embodied Dexterous Grasping System (EDGS), designed to tackle object grasping in cluttered environments for human-robot interaction. We propose a novel approach to semantic-object alignment using a Vision-Language Model (VLM) that fuses voice commands and visual information, significantly enhancing the alignment of multi-dimensional attributes of target objects in complex scenarios. Inspired by human hand-object interactions, we develop a robust, precise, and efficient grasping strategy, incorporating principles like the thumb-object axis, multi-finger wrapping, and fingertip interaction with an object's contact mechanics. We also design experiments to assess Referring Expression Representation Enrichment (RERE) in referring expression segmentation, demonstrating that our system accurately detects and matches referring expressions. Extensive experiments confirm that EDGS can effectively handle complex grasping tasks, achieving stability and high success rates, highlighting its potential for further development in the field of Embodied AI.
    \end{abstract}

    
    \begin{figure*}[t]
      \centering
      \includegraphics[width=\textwidth]{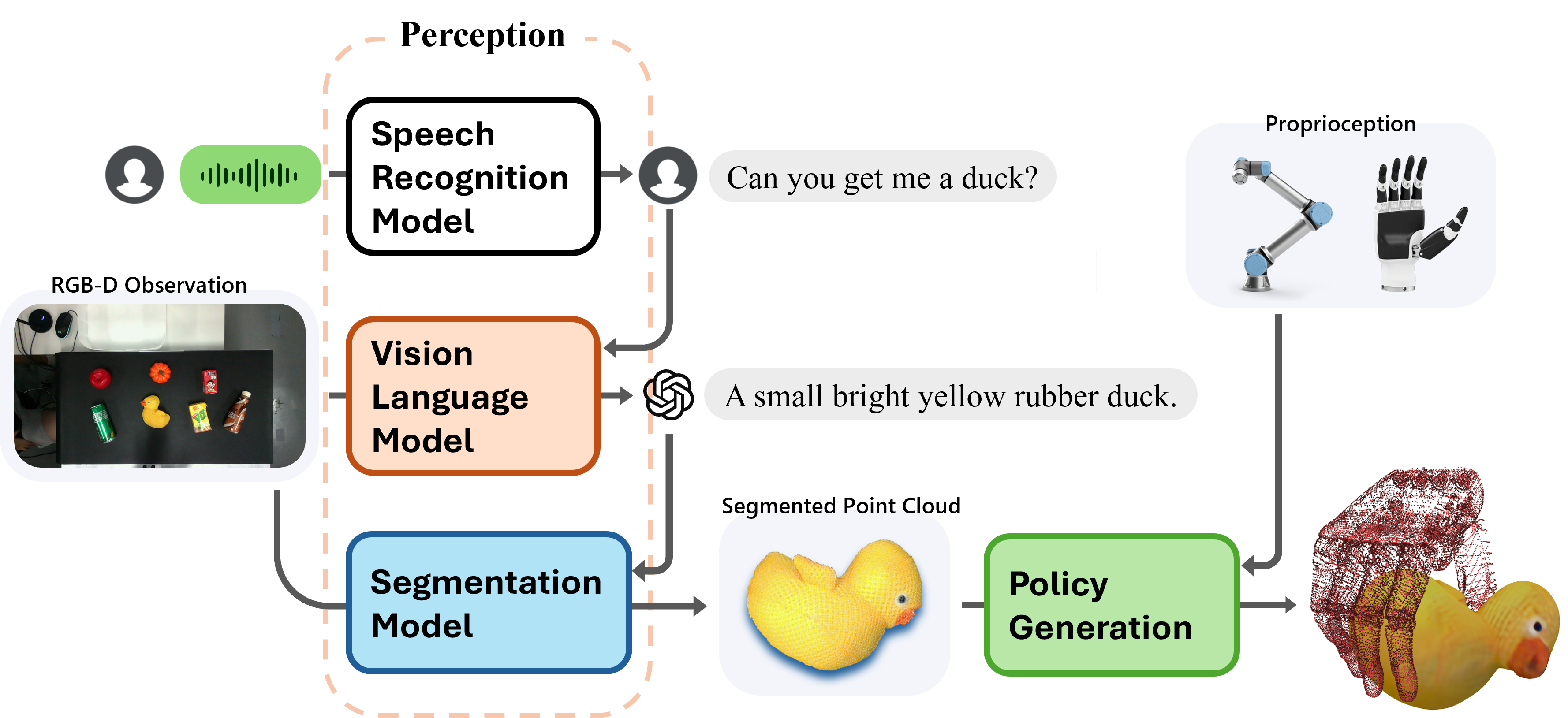}
      \caption{\textbf{Overview of Our Method.} The system processes natural language instructions via a speech recognition module and RGB-D scene data with a vision-language model to generate a r ich object description. A segmentation model isolates the target object, creating a segmented point cloud. The policy generation module then computes grasp strategies, executed by a robotic arm with proprioceptive feedback for precise manipulation.}
      \label{fig:method_overview}
    \end{figure*}
    
    \section{Introduction}
    \label{sec:intro}
      In recent years, advancements in robotics have increasingly focused on enhancing the collaboration between humans and machines, with an emphasis on intuitive control of the robotic systems. The research and development of embodied dexterous grasping systems driven by voice commands represent a critical branch in advancing this field. By integrating voice instructions with visual information, such systems enable robots to better comprehend human intentions, execute complex reasoning and decision-making processes, and perform sophisticated motion planning. These capabilities empower robots to complete dexterous grasping tasks in intricate and dynamic environments, bridging the gap between human-robot interaction and practical deployment. 
    
      Grasping target objects in cluttered and unstructured environments based on human instructions is a core challenge for robots. The capability of robotic systems to perform dexterous grasping tasks driven by imprecise natural language commands has become increasingly critical. These tasks not only require robots to understand diverse instructions but also to leverage the multi-dimensional properties of objects and dynamic adaptation during the grasping process. 
    
      Current dexterous grasping solutions can be divided into two main categories: data-driven methods \cite{DBLP:conf/icra/MosbachB24, lum2024dextrah, 10493125} and analytical approaches \cite{dingopen6dor, li2024geometric}. Data-driven methods learn grasping strategies from large-scale datasets, offering strong generality to various objects and scenes. However, these methods face major challenges in real-world deployment due to high computational costs, the lack of datasets for dexterous hands, and poor simulation-to-reality transfer \cite{huang2024rekep}. On the other hand, while analytical methods excel in controlled environments, they often struggle to accommodate vague human instructions in complex, cluttered scenes. Additionally, some of these methods rely on pre-existing CAD models, which restricts their generalization capability. \cite{hinterstoisser2011gradient, lin2024sam}.
        
      To address the limitations of scenario generality and deployment accuracy in cluttered scenes, we propose \textbf{EDGS}, a modular framework designed to optimize the dexterous grasping process by integrating structurally enriched language instructions with a comprehensive grasp policy. Our system allows users to initiate instructions through oral language, offering more direct and practical interaction that aligns seamlessly with real-world scenarios. In human cooperation, individuals continuously align their cognitive understanding with external environmental cues to achieve more precise descriptions, thereby reducing ambiguity. Inspired by this intuition, we introduce RERE, which leverages large vision-language model\cite{achiam2023gpt} to align specified objects in images with instructions across modalities. This approach enhances the referring expression representation of the target object by leveraging the model's knowledge, enabling more precise semantic-object alignment.
    
      In addition to advancement in perception level, we propose a dexterous grasping policy that leverages the hand's natural proficiency in adaptability to various objects, which outperforms conventional parallel grippers\cite{casas2024multigrippergrasp}. Biologically, humans instinctively leverage their proficiency in thumb-fingers coordination to optimize their grasping\cite{cutkosky1990human}. They perceptually locate the hand-object intersection that, from their predictive assumption, easiest for the object to be held and grasped by the coordinated movement of the thumb and other fingers.
    
      Building on this insight, we propose a Dexterous Grasp Candidates Generation (DGCG) method, which systematically performs a skeleton-based object feature extraction, and utilize the extracted feature to perform constrictive grasp action sampling strategy.
      
      This approach, inspired by human behavioral studies, robustly sample the potentially successful grasp poses, greatly improving computational efficiency and ensuring physical feasibility. 
    
      Building upon the generated grasp candidates, our Dexterous Grasp Refinement (DGR) module evaluates and optimizes grasp action candidates to ensure robust and efficient execution. This process begins with a force-closure validation, which filters grasps by assessing their ability to resist external forces through stable contact point configuration. To further quantify the robustness, we employ Grasp Wrench Space (GWS) \cite{1307170} analysis, leveraging the convex hull of frictional forces and torques to evaluate the resistance to perturbations, selecting grasps with superior quality metrics. Finally, inspired by the principle of minimizing human effort in movement, we incorporate a motion cost evaluation, utilizing inverse kinematics and trajectory planning to identify the grasp with the shortest and most efficient path.
    
    Our contributions are as follows:
    1) We construct RERE method, enhancing semantic-object alignment in cluttered environments by enriching instruction using vision-language model;
    2) We design a human-behavior-inspired robust dexterous grasping strategy integrating DGCG and DGR, enhancing both effectiveness and efficiency by applying highly constrained candidate sampling strategy and physically grounded refinement for stable and precise grasps;
    3) Extensive real-world experiments demonstrate that our system achieves high success rates and robust performance in diverse and cluttered grasping tasks.

    
    \section{Related Work} 
    \label{sec:related_work} 
    
    \vspace{-5mm}
    \subsection{Referring Expression Representation} 
    Referring Expression Representation (RER) plays a pivotal role in robotic perception, particularly in ensuring accurate target detection and localization. 
    In cluttered environments, the sufficiency of target descriptions in RER directly enhance detection performance. 
    Insufficient human instruction often make it challenging to distinguish the target object, thereby reducing grasping success rates. 
    Consequently, refining RER to minimize linguistic and contextual ambiguities has become a central focus in recent research.
    
    The advancement of Referring Expression Segmentation (RES) has significantly enhanced the application of RER in robotic operations. Methods such as GRES \cite{GRES} have extended RER by incorporating multi-target expressions, implicitly modeling spatial relationships among objects to achieve effective segmentation in complex  scenes. Similarly, LAVT \cite{yang2022lavt} introduces language features during the visual encoding stage, enabling improved cross-modal alignment and enhanced object recognition accuracy. Furthermore, Text4Seg \cite{lan2024text4seg} redefines segmentation as a text generation task, leveraging textual representations of images to improve segmentation precision.
    
    Progress in segmentation techniques has further expanded RES capabilities. Grounded SAM \cite{ren2024grounded} integrates transformer-based language models to detect and segment objects, and its segmentation accuracy mostly depends on the accuracy of human instructions. SEEM \cite{zou2024segment} explores fine-grained object-level features for multimodal semantic segmentation. Florence-2 \cite{xiao2024florence} employs a sequence-to-sequence framework to combine image embeddings with textual descriptions, generating high-quality segmentation masks and performing robustly even under ambiguous instructions.
    
    The evolution of multimodal representation learning is vital in RES development. A survey on multimodal machine learning \cite{baltruvsaitis2018multimodal} underscore the importance of integrating diverse modalities to enrich contextual understanding. Approaches such as multimodal deep learning \cite{ngiam2011multimodal} and self-corrected multimodal large language models \cite{liu2024self} further highlight the potential of leveraging multimodal data to improve downstream tasks, including segmentation and grasp planning.
    
    To address the limitations caused by the inherent vagueness of instinctive human instruction, we propose RERE, a novel module which enriches target descriptions with attributes like  shape, color, texture, and material. By cross-modally aligning linguistic and visual information, RERE reduces semantic ambiguity, improves intent-object matching, and boosts performance in recognition, localization, and grasping in cluttered environments.
    
    \subsection{Dexterous Grasping System}
    
    Significant efforts in robotic grasping focus on improving success rates across diverse objects, particularly in cluttered scenes. Geometric-based methods, such as point clouds \cite{pas2017graspposedetectionpoint}, geometric decomposition \cite{li2024shapegraspzeroshottaskorientedgrasping}, and instance segmentation \cite{10493125}, have been widely used to estimate grasp poses. For instance, Li et al. \cite{li2024shapegraspzeroshottaskorientedgrasping} decompose objects into simpler convex shapes and represent them with graph structures to model spatial relationships, while Pas et al. \cite{pas2017graspposedetectionpoint} employ point cloud processing to select optimal grasp poses. These approaches enhance grasp estimation by leveraging geometric understanding, especially in complex environments.
    
    To improve generalization to objects of various shapes, color, textures, and materials, data-driven approaches have been adopted. Some studies, like Wang et al. \cite{wang2023dexgraspnetlargescaleroboticdexterous}, generate large-scale synthetic datasets using depth-accelerated force closure estimators to synthesize stable grasps, while Vuong et al. \cite{vuong2023graspanythinglargescalegraspdataset} employ text-to-image generation and pre-trained models to develop grasp policies from diverse synthetic images. However, the gap between synthetic data and real-world conditions hinders the deployment of these methods. In contrast, more integrated approaches aim to close the sim-to-real gap. GraspGPT \cite{10265134} enhances task-oriented grasping through semantic-driven zero-shot generalization, and AnyGrasp \cite{10167687} trains grasp perception models on real-world data to generate dense and accurate grasp poses in dynamic scenes. Despite these advances, most of these approaches are limited to parallel-jaw grippers, lacking the adaptability and dexterity needed for more complex manipulation tasks involving multi-fingered hands. Wang et al.\cite{wang2024neuralattentionfieldemerging} propose a Transformer-based neural attention field method that leverages self-supervised learning to enable one-shot generalization of dexterous grasping across varying scenes, enhancing semantic awareness and robustness in grasping tasks.
    
    Reinforcement learning (RL) offers distinct benefits for handling high-dimensional inputs and dynamic environments in robotic grasping. Mosbach et al. \cite{DBLP:conf/icra/MosbachB24} introduce the Teacher-Augmented Policy Gradient framework, combining policy distillation with RL to enable zero-shot transfer from simulation to real-world grasping. Another approach \cite{9561802} integrate object-centric visual affordances into a deep RL framework, guiding robots to learn stable, functional grasp strategies. Lum et al. \cite{lum2024dextrah} combine reinforcement learning, geometric fabrics, and teacher-student distillation to train a depth-based dexterous grasping policy (DextrAH-G), but it can only handle one object in the scene at a time. Although RL excels in adaptability, its generalization and performance in cluttered scene remains a challenge, often leading to failures when encountering unseen objects. It excels in using a dexterous hand to perform grasping tasks, but prior information on single objects remains essential.

    Some hybrid methods attempt to combine RL with data-driven techniques to improve generalization and sim-to-real transfer \cite{10205189,10378503}. However, these methods still struggle with real-world deployment due to friction mismatch and contact force instability. In contrast, our approach demonstrates a notable advantage by a  robust grasp policy, achieving a higher grasp success rate in challenging cluttered scenes.
    
    \section{Method}
    
    In this section, we introduce the EDGS framework, which integrates multimodal instruction processing with a robust grasp planning and optimization pipeline, specifically designed for dexterous robotic hands. Our approach enables precise, adaptive grasping across diverse object types and cluttered environments. As shown in Figure \ref{fig:method_overview} our method comprises three primary modules: Enriched Representation Guided Segmentation (ERGS), DGCG, and DGR.

    \subsection{Enriched Representation Guided Segmentation}
    In complex environments, the effective integration of textual descriptions and visual information is crucial for accurate object recognition. Traditional methods often rely solely on textual input for object identification, but in many cases, language descriptions may be ambiguous or incomplete, leading to misinterpretations. In contrast, humans frequently use visual cues in daily interactions to resolve ambiguity in language, thereby improving the precision of task execution. Inspired by this phenomenon, we propose a novel approach RERE which enhances textual descriptions through enriched visual information to improve object segmentation accuracy.

    The core idea of RERE is to refine and supplement the user's original language input with visual data, reducing uncertainty in the textual description. This process depends on a cross-modal alignment mechanism that combines visual cues with textual information, helping to clarify and correct ambiguities. For instance, when a user says "Give me a duck on the table, please.", visual information can help the system clarify the type, color, material, and spatial position of the object, thus effectively filtering out interference from similar objects. Previous studies have shown that reducing the ambiguity inherent in textual inputs can significantly enhance a model’s capacity to identify and interact with objects \cite{lu2024evaluation}. 
    
    \vspace{-3mm}
    \paragraph{Voice-Driven Referring Expression Acquisition}

    The segmentation process in RERE begins with acquiring an initial description of the target object via speech commands. However, speech input is often affected by noise and linguistic ambiguity, which can distort or degrade the accuracy of the information. To address this, we employ an adaptive audio capture method, using signal energy  \(E(t)\) to dynamically set a threshold for audio capture. The capture is triggered only when the signal energy exceeds the threshold, and stops when it remains below the threshold for an extended period. This strategy helps avoid redundant or erroneous audio data.

    Assuming that the original audio signal \(A(t)\) has been transcribed into text \( T_{\text{orig}} \), this text represents the user's natural language input, but it is often vague and may affect the accuracy of subsequent object segmentation. To resolve this ambiguity, RERE introduces a multi-modal fusion mechanism that combines visual information with the text to enhance its clarity and precision.

    \vspace{-3mm}
    \paragraph{Referring Expression Representation Enrichment}
    
    A critical step in RERE is the cross-modal alignment process, where the system evaluates the alignment between the original text description \( T_{\text{orig}} \) and the visual input \( I \in \mathbb{R}^{h \times w \times 3}\). If the alignment is deemed sufficient, the system proceeds with the task; otherwise, it prompts the user for clarification, ensuring that only well-aligned inputs contribute to accurate object identification.

    To enhance object segmentation further, we have designed a rule-based system to semantically enrich the original text based on various visual features. This process reduces the ambiguity in the text and ensures that the description of the target object is more precise. Specifically, RERE enhances the semantic representation of the target object along the following key dimensions: 1) Instance Category: By analyzing visual cues, the system identifies the type of the target object, ensuring that the description is consistent with the visual scene, particularly in multi-object environments; 2) Color and Shape: Color and geometric features extracted from the visual input are incorporated to provide a clearer and more accurate description of the object’s visual properties; 3) Material and Texture: The system identifies unique surface characteristics (e.g., metal, glass) to highlight distinguishing material features, aiding in object differentiation; 4) Position: The spatial position of the object within the scene (e.g., "on the left," "at the center") is integrated into the description to further specify the object’s location, reducing ambiguity from surrounding elements.

    The final enriched text description \( T_{\text{final}} \) is generated through a function that integrates the original text description  \( T_{\text{orig}} \) , visual features, and contextual cues. This process can be represented as:
    
    \begin{equation}
    \resizebox{\columnwidth}{!}{$T_{\text{final}} = f(T_{\text{orig}}, \mathcal{A}(T_{\text{orig}}, I), V_{\text{feat}}(I), \mathcal{C}(T_{\text{orig}}, I), w_C, w_S, w_M, w_P)$}
    \end{equation}

    where:
    \begin{itemize}
    \item $T_{\text{final}}$
    is the enhanced textual description, which provides a semantically clearer and more accurate representation of the object. 
    \item $f(\cdot)$ is a mapping function that generates the final description by combining the original text, cross-modal alignment, visual features, and contextual information.
    \item $\mathcal{A}(T_{\text{orig}}, I)$ measures the alignment between the original text description and the visual input, ensuring consistency between the two modalities.
    \item $V_{\text{feat}}(I) = [C, S, M, P]$  represents the extracted visual features, including color, shape, material, and position.
    \item $\mathcal{C}(T_{\text{orig}}, I)$ incorporates contextual information that aids in refining the description.
    \item $w_C, w_S, w_M, w_P$ 
    are weight parameters that allow the system to adjust the relative importance of color, shape, material, and position features based on the scene.
    \end{itemize}

    For vague or incomplete user inputs, RERE utilizes contextual information to guide the generation of the most appropriate description, rather than relying solely on model inference. This process leverages contextual cues from the scene to optimize imprecise descriptions, ensuring that each description provides sufficient information for subsequent tasks. For example, when the user inputs "the red object," the system can refine the description to specify whether it is "a red circular cup" or "a red square box," based on the visual input.
    
    Moreover, RERE emphasizes the importance of multi-modal information fusion. This fusion is not merely about concatenating visual and linguistic features; rather, it dynamically adjusts the weight assigned to different modalities based on the specific requirements of the scene, selecting the most distinguishing or salient features for representation. This dynamic adjustment mechanism ensures that the model remains robust and efficient in complex environments.

    In summary, RERE significantly enhances object segmentation accuracy by guiding the enrichment of textual descriptions with visual information within a multi-modal learning framework. As highlighted in research on multi-modal learning, the integration of different modalities enhances the model's expressive power and robustness \cite{baltruvsaitis2018multimodal, ngiam2011multimodal}.

    \begin{figure*}[ht] 
      \centering 
      \includegraphics[width=\textwidth]{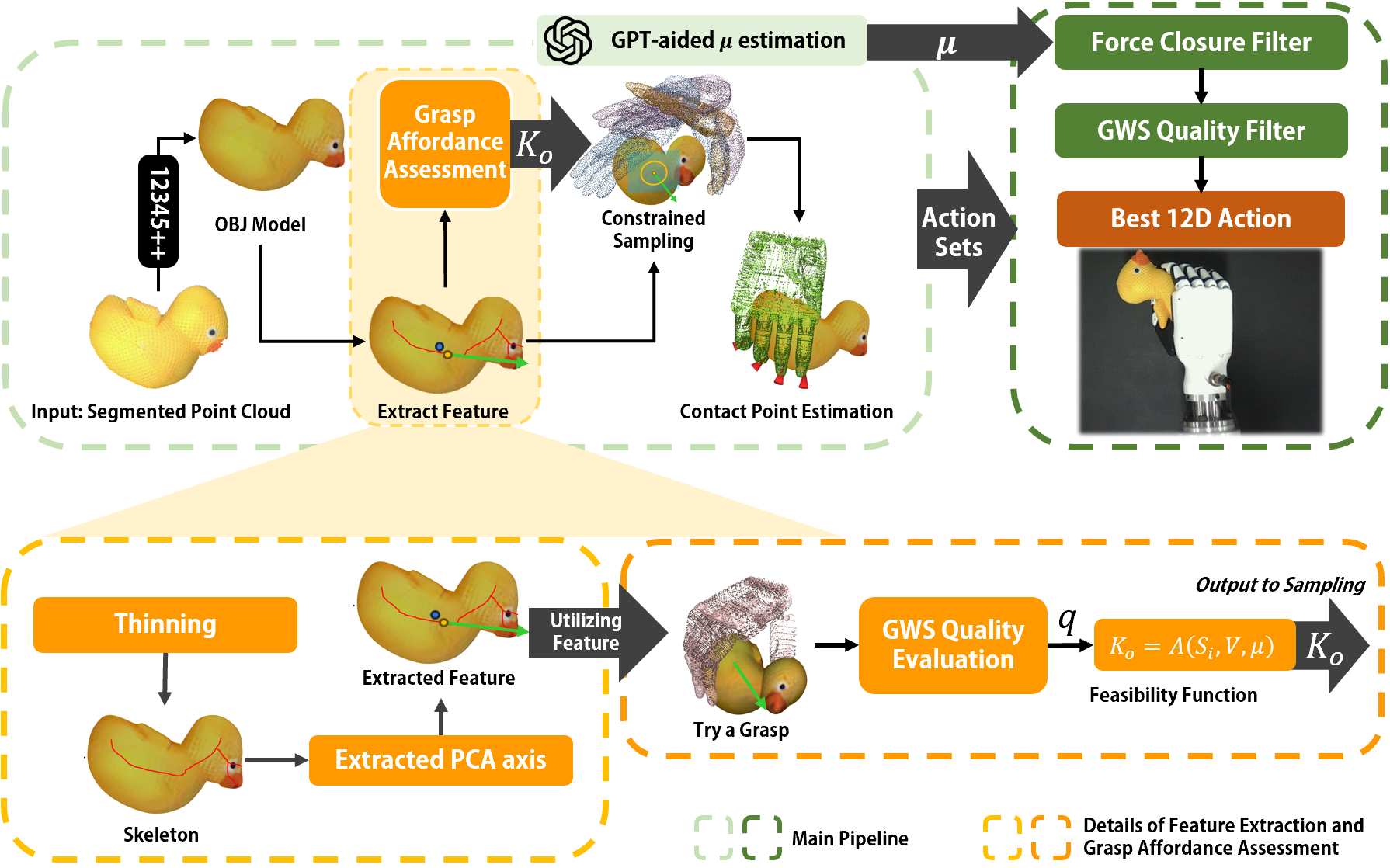}
    
      \caption{\textbf{Grasp Policy Generation Method.} Our method starts with segmented point cloud for feature extraction, followed by constrained sampling and contact point estimation. Parameter (\( K_o \)) related to constrained sampling is determined by grasp affordance assessment. A GPT-aided module estimates the friction coefficient (\( \mu \)) for force closure filtering, and the grasp action sets are refined through GWS quality assessment to determine the best 12D action. } 
      \label{fig:grasp_policy_overview} 
    \end{figure*} 
    
    \subsection{Dexterous Grasp  Candidates Generation}
    
    Given a segmented point cloud data , we perform object feature vector extraction, based on which we generate a set of action candidates. 
    As shown in Figure 
    \ref{fig:grasp_policy_overview}, our Grasp Candidates Generation unfolds in the following three phases: 1) Object Feature Extraction; 2) Grasp Affordance and Candidates Generation
    
    \vspace{-3mm}
    \paragraph{Object Feature Extraction} 
    
     When humans perform grasping tasks, they instinctively adjust their hand pose and wrist position and orientation. This ability stems from the integration of multiple sensory inputs of an object's position and orientation in complex environments\cite{Betti2021ReachToGrasp}. Inspired from this idea, we develop the following mechanism to process the geometric information from RGB-D data, determining the characteristic centeroid and subsequently the feature vector. 

     Merely extracting Principal Component Axis has inherent limitations for unsymmetrical  objects to derive a main axis, that's why it's required to extract and process the object skeleton line to represent its orientation\cite{blum1967transformation}. Especially for unsymmetrical , irregularly shaped , morphologically-complicated objects, we need to use skeleton lines combined with PCA to represent the morphological orientation and extension.  We extract the skeleton of the segmented object using thinning algorithm \cite{zhang1984fast}, and from this line, we define the point closest to the pixel centroid \(C\) as the characteristic centroid \(C^*\).

    \begin{equation}
        C^* = \arg\min_{(x_e, y_e) \in E} \sqrt{(x_e - C_x)^2 + (y_e - C_y)^2}
    \end{equation}
    where  $ E = \{ (x_e, y_e) \mid e = 1, 2, \dots, N \} $, as the set of skeleton points. 
    
    We use the skeleton lines in 2D space, with the tangential line at the C* points, as the feature vector in the X-Y plane direction. To minimize the impact of noisy depth data, we leverage the advantage of PCA axis\cite{10.1145/1618452.1618522}. The eigenvector $\mathbf{v}_1$ corresponding to the largest eigenvalue defines the direction of the Principal Axis $V_1$. We assign this axis on characteristic centroid \(C^*\)  , and accordingly define this vector starting from \(C^*\) as an object feature vector. 

    \begin{equation}
        V = \begin{pmatrix} 
            C^*_x + t T_{x}, & C^*_y + t T_{y}, & C^*_z + t V_{1z} 
        \end{pmatrix}
    \end{equation}

    \paragraph{Grasp Affordance and Candidates Generation}

    In this paper, we propose a novel approach for evaluating the feasibility of grasps by integrating the object feature vector with the hand configuration. We model the workspace of each finger as a truncated spline, derived from the movement of the fingertip. Given that each finger has a single degree of freedom at the distal joints, the workspace of the fingers is represented by a set of truncated splines:
    
    \begin{equation}
    W_{\text{finger}} = \{ W_1, W_2, W_3, W_4 \},
    \end{equation}
    where each \( W_i \) corresponds to a specific truncated spline for the \(i\)-th finger. The thumb, having two degrees of freedom, defines a planar workspace \( W_{\text{thumb}} = P_o \), where \( P_o \) represents a planar space. Combining the workspaces of the fingers and thumb, the total workspace \( W_{\text{total}} \) is given by the union of all individual workspaces: 
    \begin{equation}
    W_{\text{total}} = W_{\text{finger}} \cup W_{\text{thumb}}.
    \end{equation}
    Inspired by the principles of human grasping behavior \cite{bongers2012hand}, and utilizing the URDF of the dexterous hand, we establish a fixed pinching axis \( \mathcal{D} \) within the hand's workspace. This axis, based on the observed coordination between the thumb and the other four fingers, is aligned with the object feature vector to define the pinching direction of the hand. The hand's workspace is then projected into space at a fixed tilt angle, allowing us to compute the intersection with the object model. To evaluate the quality of a given grasp, we use the GWS mechanism, as detailed in Section \ref{dexterous_grasp_refinement}. 
    
    The hand's workspace is then projected on this hand root position, to obtain the intersections \(S\) between the hand's workspace \(W_{total}\) and object surface \(O\) will be determined by the following equations: $ S_i = W_{total} \cap O  $ , which geometrically consists of four intersected points  \(I_1,I_2,I_3,I_4\) corresponding to non-thumb fingers, and one curve \(L_{thumb}\) corresponding to the set of all possible touching points of the thumb. 

    For grasp planning, we define an affordance function \( \mathcal{A}(S_i, V, \mu) \) to describe the probability of a given grasp configuration. This function is formulated as:

    \begin{equation}
        K_o = \mathcal{A}(S_i, V, \mu) = \Phi(f_w(S_i, V, \mathbf{\mu}))
    \end{equation}

    where \( \mu \) is the frictional coefficient returned by vision language model. The function \( f_{w}(S_i, V, \mathcal{\mu} ) \) is a mapping that evaluates the grasp quality based on the position and orientation (utilizing the workspace constructed at the current position and rotations to obtain intersection) , and \( \Phi \) is a normalization function that returns the affordance score for a given grasp configuration. This approach provides a comprehensive framework for grasp affordance evaluation by considering both the spatial position and orientation of the grasp, thus enabling a more nuanced assessment of graspability.

    Furthermore, the affordance score is used to define the range of this Gaussian distribution, and is processed accordingly : 
    {\small
    \begin{equation}
    G(m) = \frac{1}{\sqrt{(2\pi)^{k}|\Sigma|}} \exp\left(-\frac{1}{2} (m - \mathcal{D})^T \Sigma^{-1} (m - \mathcal{D}) \right)
    \end{equation}
    }

    where \( m \in \mathbb{R}^6 \), the dimension \( k = \)6, capturing both spatial and rotational aspects of the grasp, and \( |\Sigma| = \frac{1}{K_o} \times I\) is the covariance matrix.

    For each \( m_i \) in the set \({\{M_i\}^N}\),we compute the corresponding object-hand intersection \(S_i\), and apply inverse kinematics to determine the hand joint space configuration \(h_i \in \mathbb{R}^6\) of each grasp. Then \(\{H_i\}^N\) is concatenated with \(\{M_i\}^N\), to construct the set of our generated action candidates \(\{A_i\}^N\).

    \subsection{Dexterous Grasp Refinement}
    \label{dexterous_grasp_refinement}
    
    In this section, grasp candidates $ \{A_i\}^{N} $ are discriminated by the following filter mechanisms to eventually determine the best Action result $ A_i^{best} $ . The quality of each sampled grasp is, inspired by the mechanism proposed by Wang et al.\cite{chen2023task} , evaluated based on both Force Closure and GWS.
    
    \vspace{-3mm}
    \paragraph{Grasp Quality Evaluation}
    
    Humans not only estimate properties like mass and surface friction through visual and tactile modalities but also intuitively adjust their grasping postures and strategies based on these perceptions\cite{DBLP:journals/corr/abs-2112-14119}. This rough, granular assessment forms the basis for adaptive interactions with objects, often relying on prior knowledge and classification of texture. 
    
    In this context, \( \mu_i \) is derived from ground truth values in a priori tables, informed by texture classification through VLM.  Force Closure\cite{1087483} determines the success of a grasp pose, while GWS \cite{1307170} and its spatial tensors quantitatively evaluate grasp quality by assessing the resistance to external disturbances, such as instantaneous forces and moments.
    
    \begin{equation}
    \exists S \left( \bigcup_{d \in S} C(d_j) \neq \emptyset \right)  \Rightarrow \text{Force Closure}
    \end{equation}
    where \( C(d) \in F\left(\arctan(\mu_i), F_i\right)\), \(d_j\) refers to every point extracted from surface \(S\) . 
    
    Grasp Candidates satisfying force closure could then be selected to form a smaller Action set  $\{A_i\}^{n} \in \mathbb{R}^{12}$ .
    
    GWS is defined as the set of all frictional forces and pressures combined: 
    {\small 
    \begin{equation}
    GWS = Conv \left( \{ W = \begin{bmatrix}
    F_x & F_y & F_z \\
    \tau_x & \tau_y & \tau_z
    \end{bmatrix} \mid \|w\| \leq 1 \} \right)
    \end{equation}}
    Where , $w_i = \sum_{j=1}^{n} \lambda_j w_j$, $\lambda_j \geq 0$, and $\sum_{j=1}^{n} \lambda_j = 1$. Centered at the geometric center of this convexhull, the radius of the smallest 6D sphere  \(q\)  that fully encloses the GWS captures the grasp’s disturbance resistance capability . Besides, object-hand intersected area \(S_i\) could be converted to the possible solution space of local normal forces in Orthogonal Decomposition, based on frictional cone theory. 
    
    For $  \|w\| \leq r ), \ p_i \in \mathbb{R}^{12} $, 
    we select the top three grasps $\{A_i\}^{3} \in \mathbb{R}^{12}$ with the highest \(\mathbf{q}\) values. 
    {\small
    \begin{multline}
        \mathcal{Q} = \left\{ 
        (q_i, p_i) \ \middle|\ q_i = \max \left( r \ \middle|\ \forall w \in \text{ConvHull} \left( \{ (F_j, \tau_j) \} \right) \right) 
        \right\}, \\
        i = 1, 2, 3.
    \end{multline}
    }
    
    \vspace{-3mm}
    \paragraph{Cost-based Movement Selection} In human motion, there is an inherent tendency to select the optimal movement, directly reaching the target with minimal effort, avoiding unnecessary muscle tension and energy expenditure\cite{umberger2017optimal}. We employed the Stochastic Trajectory Optimization for Motion Planning (STOMP) method to optimize paths and identify the optimal, smooth trajectory in complex environments. To further enhance the rationality and optimality of trajectory planning, the following path constraints were incorporated: 1) Pose Constraints: Ensuring that the end-effector maintains alignment with the target object throughout the grasping process; 2) Joint Constraints: Restricting joint movements within their permissible limits to ensure the feasibility and safety of the planned trajectory; 3) Dynamic Obstacle Avoidance: Utilizing the obstacle avoidance capabilities of MoveIt to dynamically adjust the trajectory and prevent collisions with obstacles.By integrating these constraints, we achieved optimal trajectory planning tailored to the best grasping pose, significantly improving the success rate and execution efficiency of grasping tasks.Then we select the optimal grasp as follows:
    
    \begin{equation}
    A_i^{best} \xleftarrow{} \arg\min \left( \sqrt{\sum_{k=1}^{n} \left( \|\Delta J_{i,k}\|^2 \right)} \right)
    \end{equation}
    
    where \(\Delta J_{i,k}\) represents the 6D Joint difference calculated from Inverse Kinematics. 
    
    \section{Experiment}
    \label{experiment}
    
    Following the setup description, we present the experimental evaluation of EDGS, which includes a detailed investigation of its performance in RERE method, grasp success rates, and real-world application scenarios. 
    
    \subsection{Set Up}

    \begin{figure}[ht]
    \centering
    \subfloat[Single-Arm Dexterous Grasping Platform]{
        \begin{minipage}{0.9\linewidth}
            \centering
            \includegraphics[width=1.0\linewidth]{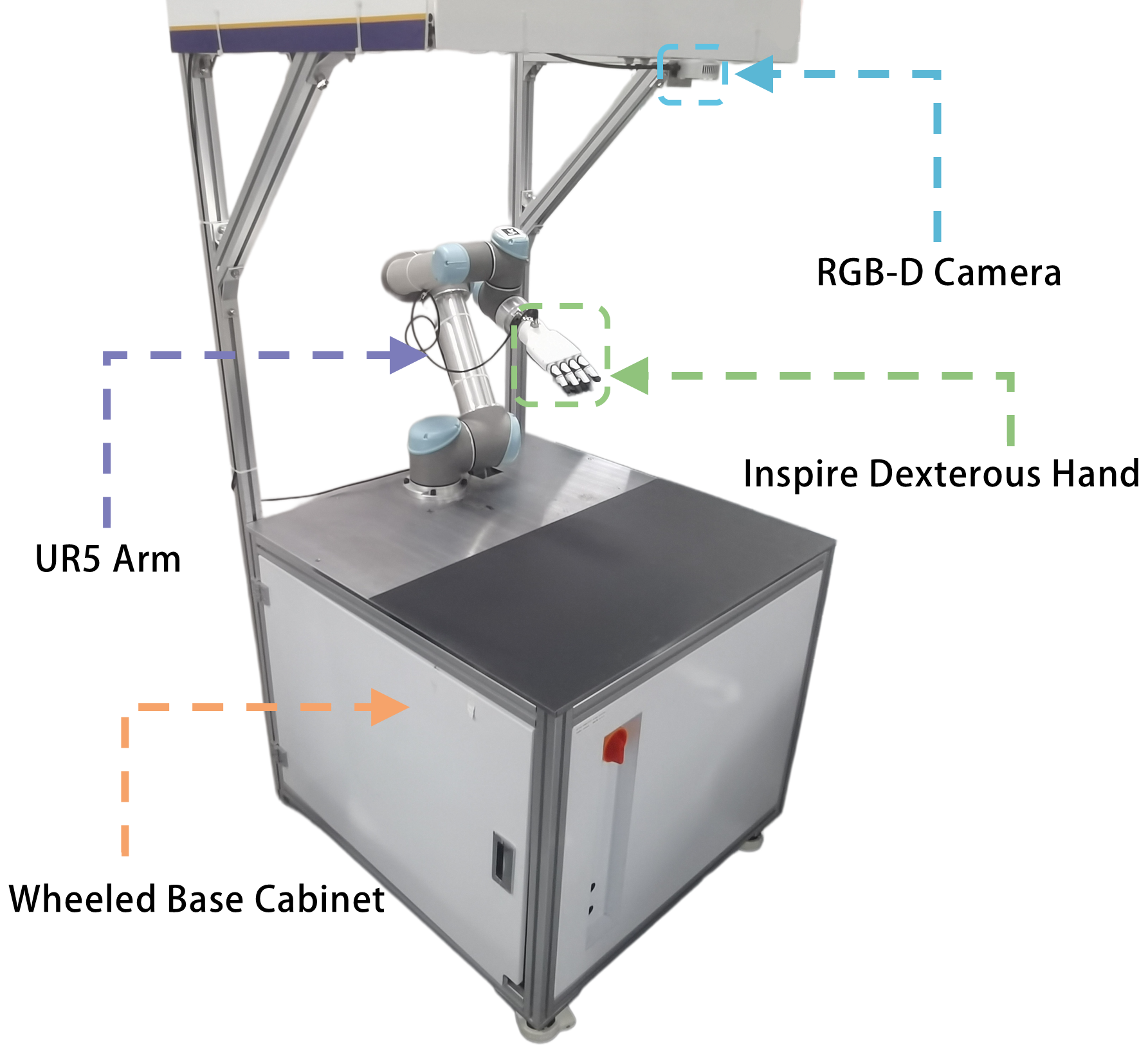} 
        \end{minipage}
    }
    \hspace{0.5\linewidth}
    \subfloat[Objects used in single and object-by-object grasping experiments.]{
        \begin{minipage}{1.0\linewidth}
            \centering
            \includegraphics[width=0.9\linewidth]{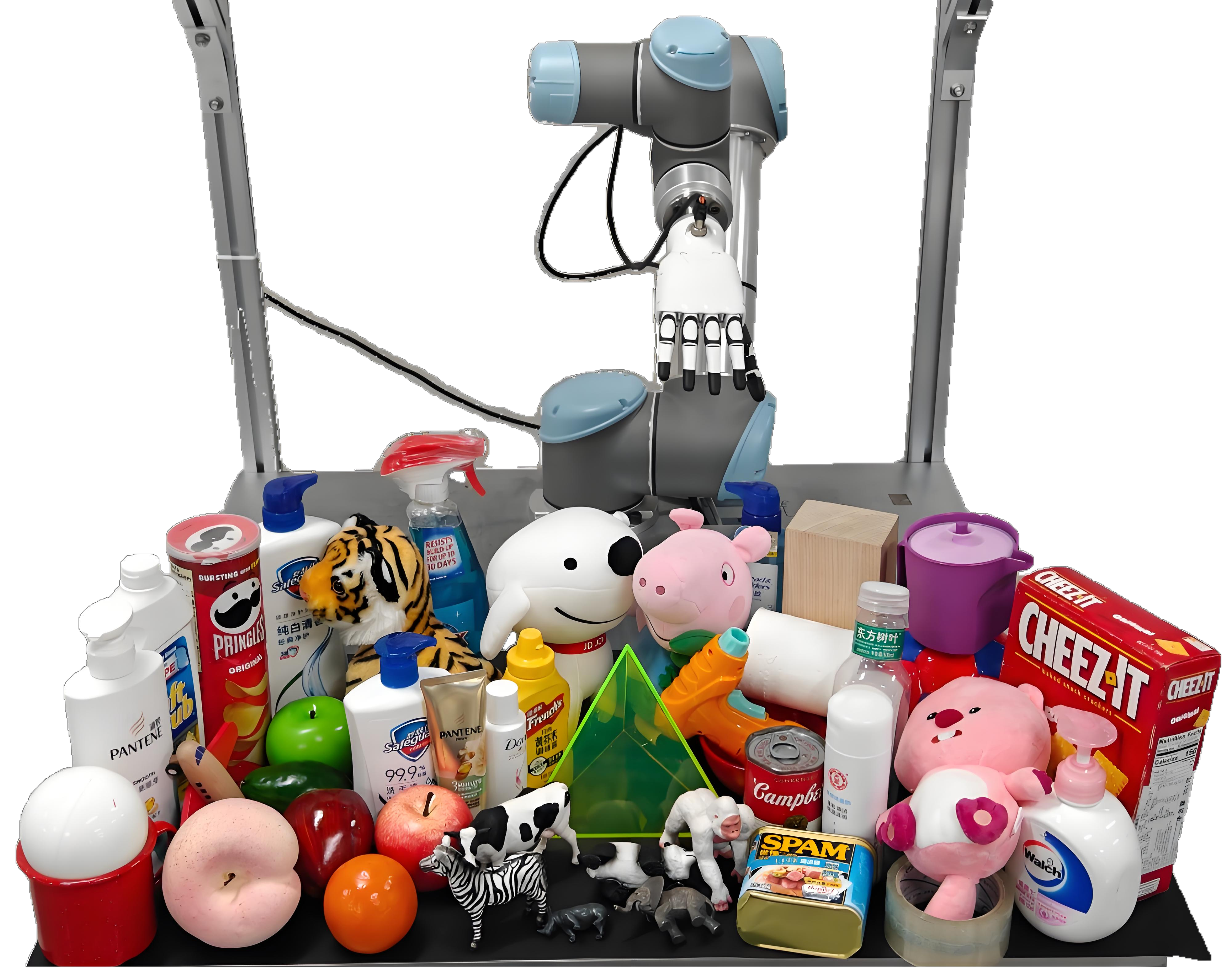} 
        \end{minipage}
    }
    \caption{\textbf{Overview of the Experimental Setup.}}
    \label{fig:The_set_up_of_experiments}
    \end{figure}
    
    We investigated this method on a single arm platform, comprising an UR5 robotic arm mounted on a cabinet (as illustrated in Figure \ref{fig:The_set_up_of_experiments}). The arm is equipped with an Inspire dexterous hand at its end. Communication between the arm and the manipulator controller is established through a high-speed Ethernet interface, ensuring the real-time responsiveness of the system. Specifically, our pipeline calculates the pose sequence required for the 6-DoF arm, which is then communicated to the manipulator. Additionally, The RGB-D camera is mounted on the upper frame of the robotic system, allowing it to capture RGB images and depth data, transmitting the data via the high-speed USB 3.1 protocol.
    
    The system can grasp a variety of household items without pre-provided CAD models. Therefore, our experiment selected various unseen household objects (as shown in Figure \ref{fig:The_set_up_of_experiments}) where we have devised subsequent experiments to thoroughly assess the robustness of the system.
    
    \subsection{Instruction semantic enrichment experiment}

    \begin{figure*}[ht]
        \centering
        \includegraphics[width=\textwidth]{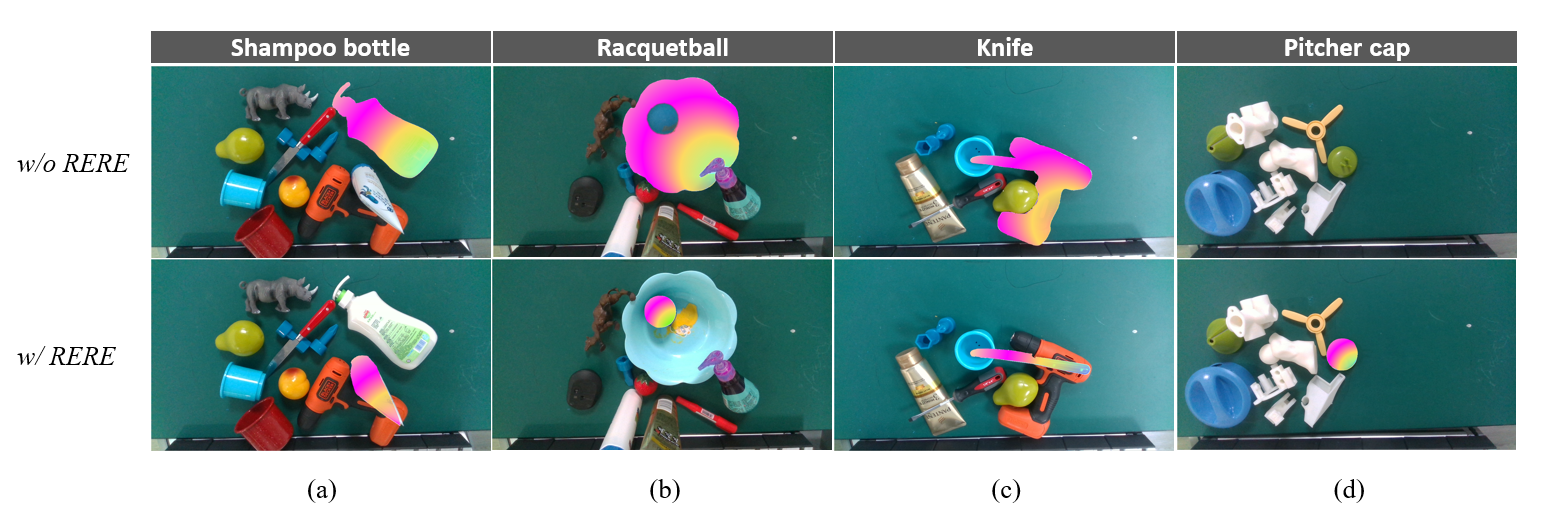}
        \caption{\textbf{Segmentation Error Analysis in Grasping Scenarios: Comparison of Results with and without RERE.} This figure shows four common segmentation errors in grasping tasks: (a) Class confusion, (b) Boundary inaccuracies, (c) Object merging, and (d) False negatives.}
        \label{fig:rere_enrich}
    \end{figure*}
    
    We conduct a series of comprehensive experiments to evaluate the effectiveness of our RERE (Refined Enriched Representation Enhancement) method in improving semantic segmentation performance across several widely-used benchmarks. Specifically, we apply the RERE technique to the GraspNet-1Billion dataset \cite{fang2020graspnet}, leveraging three distinct segmentation models to assess its impact on segmentation accuracy.
    
    The GraspNet-1Billion dataset consists of 97,280 RGB-D images spanning 88 object categories. These images are collected from 190 distinct scenes, each containing approximately ten objects, with pixel-level annotations for both objects and background across complex scene setups. Each scene has 255 different viewpoints, but for the purpose of our experiments, we reduce the diversity of viewpoints by selecting a subset that most closely resembles the real-world setup used in our experiments \cite{Zhang2024DexGraspNet2L}. This decision emphasizes the importance of scene diversity over viewpoint diversity, ensuring that the dataset remains representative of the types of challenges encountered in practical semantic segmentation tasks.
    
    Our proposed method enhances the original caption descriptions by incorporating information from raw captions, RGB images, and a carefully designed prompt. The prompt guides the large language model (LLM) to generate enriched captions from multiple dimensions. These enriched captions are then employed to generate object masks using state-of-the-art segmentation models, including Grounded SAM \cite{ren2024grounded}, SEEM \cite{zou2024segment}, and Florence-2 \cite{xiao2024florence}. The segmentation performance is quantitatively assessed by comparing the mean Intersection over Union (mIoU) of the generated masks with the ground truth annotations.
    
    The experimental results, summarized in Table \ref{tab:method_comparison}, clearly demonstrate that the RERE method significantly improves segmentation performance across all three models. Specifically, improvements in IoU range from 5.0 to 25.7 points. Notably, Grounded SAM exhibits the most substantial enhancement, with its IoU score nearly doubling from 44.9 to 64.4, a remarkable 19.5 point increase. SEEM and Florence-2 also show notable improvements of 12.2 and 7.3 points, respectively. These results underscore the efficacy of the RERE method in boosting segmentation accuracy, particularly for models like Grounded SAM, which appears especially sensitive to the enriched input.
    
    To further explore the potential of RERE, we conduct ablation studies to identify the optimal combination of enriched dimensions. As shown in Table \ref{tab:ablation study}, the inclusion of additional dimensions such as shape, texture, and relative position systematically improves segmentation accuracy. The most significant gain occurs when all enriched dimensions (Instance Type, Texture, Shape, Material, and Position) are incorporated, resulting in an IoU of 64.4 for the Grounded SAM model. This suggests that a comprehensive set of enriched features delivers the most effective performance.
    
    Despite the clear benefits of RERE, certain challenges remain. In instances where the original referring expressions are highly unconventional or fall outside the scope of the model's knowledge base, the alignment between modalities becomes difficult, reducing the effectiveness of RERE. Furthermore, in situations involving occluded objects, RERE tends to favor identifying similar, complete objects, which may result in misclassification or incomplete segmentation. These challenges highlight areas for future research and improvement in segmentation tasks.
    
    Figure \ref{fig:rere_enrich} presents examples of common segmentation errors encountered during grasping tasks. These errors can be categorized into four types: (a) class confusion, (b) boundary inaccuracies, (c) object merging, and (d) false negatives. These categories reveal potential limitations of the RERE approach and provide valuable insights for further refinement of segmentation strategies.

    \begin{table}[ht]
    \centering
    \resizebox{\columnwidth}{!}{%
    \begin{tabular}{l c c c}
    \toprule
    \multicolumn{1}{c|}{IoU} & Grounded SAM \cite{ren2024grounded} & SEEM \cite{zou2024segment} & Florence \cite{xiao2024florence} \\
    \midrule
    \multicolumn{1}{c|}{w/o RERE} & 44.9 & 38.2 & 48.5 \\
    \multicolumn{1}{c|}{\textbf{w/ RERE}} & \textbf{64.4} & \textbf{50.4} & \textbf{55.8} \\
    \bottomrule
    \end{tabular}%
    }
    \caption{\textbf{Comparison Experiment of RERE.} This table compares the segmentation performance of different models on all scenes in GraspNet, evaluated using IoU.}
    \label{tab:method_comparison}
    \end{table}
    
    \begin{table}[ht]
    \centering
    \resizebox{0.9\columnwidth}{!}{
    \begin{tabular}{ccccc|c}
    \toprule
    \textbf{Ins.} & \textbf{Tex.} & \textbf{Sha.} & \textbf{Mat.} & \textbf{Pos.} & GraspNet \cite{fang2020graspnet} \\
    \midrule
    & & & & & 44.9 \\
    \checkmark & & & & & 48.1 \\
    \checkmark & & & \checkmark & & 50.4 \\
    \checkmark & & & & \checkmark & 54.4 \\
    \checkmark & & \checkmark & & & 57.5 \\
    \checkmark & \checkmark & & & & 62.1 \\
    \checkmark & \checkmark & \checkmark & & & 63.9 \\
    \checkmark & \checkmark & \checkmark & & \checkmark & 64.2 \\
    \checkmark & \checkmark & \checkmark & \checkmark & \checkmark & \textbf{64.4} \\
    \bottomrule
    \end{tabular}}
    \caption{\textbf{Ablation study of RERE dimensions} on GraspNet using GSAM. Columns Ins., Tex., Sha., Mat., and Pos. represent 'Instance Type,' 'Texture,' 'Shape,' 'Material,' and 'Relative Position,' respectively.}
    \label{tab:ablation study}
    \end{table}
    
    \subsection{Grasp Success Rate Experiment}
    In this section, we present a comprehensive evaluation of the grasping capabilities of the EDGS through quantitative analysis of its performance in both single-object and Object-by-Object grasping tasks. The experiments were designed to rigorously assess the system's effectiveness in real-world scenarios, focusing on its ability to interpret voice commands and execute precise grasping actions.

    \begin{figure*}[ht]
        \centering
        \includegraphics[width=\textwidth]{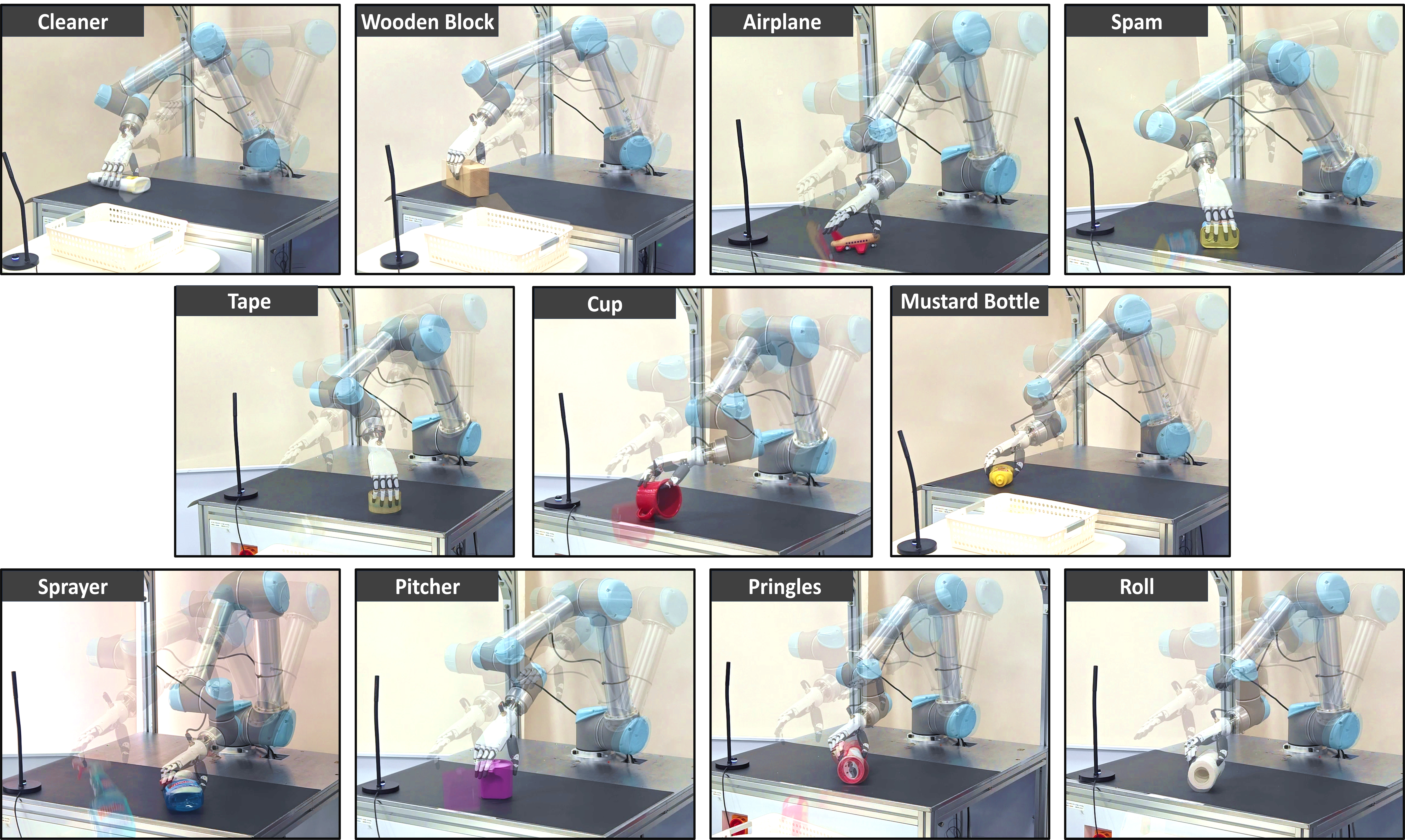}
        \caption{\textbf{Grasping Scenarios for Eleven Objects.} Experimental setups showing the system's performance across diverse objects.}
        \label{fig:grasp sequence}
    \end{figure*}
    
    \vspace{-4mm}
    \paragraph{Single Object Grasping.} In real-world scenarios, we evaluated various grasping policies by quantifying their success rates on eleven diverse objects, using a standard method that accounts for both successful grasps and the retention of the object during transfer. The success rate for each object was determined based on eight random poses, with failures defined as either the inability of the dexterous hand to grasp the object or the object slipping from the hand after a successful initial grasp. As shown in Table \ref{tab:success_rates_comparison} and illustrated in Figure \ref{fig:grasp sequence}, the EDGS method consistently outperforms other state-of-the-art approaches, achieving a 100\% success rate on objects such as the Pringles, Cleaner, Mustard Bottle, Cup, and Pitcher. Notably, EDGS demonstrates superior performance even in challenging scenarios, with an impressive 87.5\% success rate on objects like the Wooden Block and Tape, where other methods such as DexDiffuser, ISAGrasp, and Matak struggle. These results underscore EDGS’s robustness and its ability to handle complex grasping tasks, making it the most reliable approach across a variety of objects and poses.

\begin{table*}[t]
    \centering
    \resizebox{\textwidth}{!}{%
    \begin{tabular}{l c c c c c c c c c c c}
    \toprule
    \multicolumn{1}{c|}{Method} & Pringles & Roll & Tape & Cleaner & Airplane & Sprayer & Mustard Bottle & Spam & Wooden Block & Pitcher & Cup \\
    \midrule
    \multicolumn{1}{c|}{ISAGrasp \cite{Chen2022LearningRR}} & 60\% & 80\% & 80\% & - & - & - & - & - & - & - & - \\
    \multicolumn{1}{c|}{DexDiffuser \cite{weng2024dexdiffusergeneratingdexterousgrasps}} & 60\% & - & - & \textbf{100\%} & 20\% & 80\% & \textbf{100\%} & - & - & - & - \\
    \multicolumn{1}{c|}{Matak \cite{9996386}} & \textbf{100\%} & - & - & - & - & - & 35\% & \textbf{100\%} & 65\% & 67\% & - \\
    \multicolumn{1}{c|}{DextrAH-G \cite{lum2024dextrah}} & \textbf{100\%} & - & - & 100\% & 60\% & - & - & \textbf{100\%} & - & 80\% & 80\% \\
    \multicolumn{1}{c|}{EDGS(ours)} & \textbf{100\%} & \textbf{100\%} & \textbf{100\%} & \textbf{100\%} & 62.5\% & \textbf{100\%} & \textbf{100\%} & \textbf{100\%} & \textbf{100\%} & \textbf{100\%} & \textbf{100\%} \\
    \bottomrule
    \end{tabular}%
    }
    \caption{\textbf{Comparison of Grasping Success Rates Across Different Methods and Objects.} The success rates for each method are reported as percentages for each object.}
    \label{tab:success_rates_comparison}
\end{table*}

    \begin{table}[t]
        \centering
        \resizebox{0.7\columnwidth}{!}{%
        \begin{tabular}{l c}
        \toprule
        \multicolumn{1}{c|}{Metrics} & Success Rate \\
        \midrule
        \multicolumn{1}{c|}{HGCNet\cite{9811756}} & 16.4\% \\
        \multicolumn{1}{c|}{GraspTTA\cite{jiang2021graspTTA}} & 6.0\% \\
        \multicolumn{1}{c|}{ISAGrasp\cite{Chen2022LearningRR}} & 54.8\% \\
        \multicolumn{1}{c|}{DexGraspNet 2.0\cite{Zhang2024DexGraspNet2L}} & 90.7\% \\
        \multicolumn{1}{c|}{EDGS(ours)} & \textbf{95.5}\% \\
        \bottomrule
        \end{tabular}%
        }
        \caption{\textbf{Comparison of Success Rates in Multi-Object Grasping.} Success rates of different grasping methods evaluated in multi-object grasping scenarios involving various object arrangements.}
        \label{tab:EDGS}
    \end{table}
    
    \vspace{-3mm}
    \paragraph{Object-by-Object Grasping.} We evaluated our EDGS method in an Object-by-Object grasping task, where objects were sequentially removed from the scene, using the same object arrangements as in the DexGraspNet 2.0 study \cite{Zhang2024DexGraspNet2L} (see Figure \ref{fig:The_set_up_of_experiments}). The system’s performance was compared to other state-of-the-art approaches, including DexGraspNet 2.0, ISAGrasp, HGCNet, and GraspTTA, with success rates detailed in Table \ref{tab:EDGS}. In this setup, the success rate for each grasp was measured as the ratio of successful grasps to total attempts. EDGS outperformed all competing methods, achieving an impressive 95.5\% success rate. In contrast, DexGraspNet 2.0 achieved 90.7\%, while ISAGrasp, HGCNet, and GraspTTA showed much lower success rates of 54.8\%, 16.4\%, and 6.0\%, respectively. These results highlight EDGS's exceptional performance and its ability to adapt efficiently to varying object types and arrangements in multi-object grasping tasks, underscoring its robustness in cluttered environments.
    
    \subsection{Application Scenario Experiments}

    \begin{figure*}[ht]
      \centering
        \includegraphics[width=0.9\textwidth]{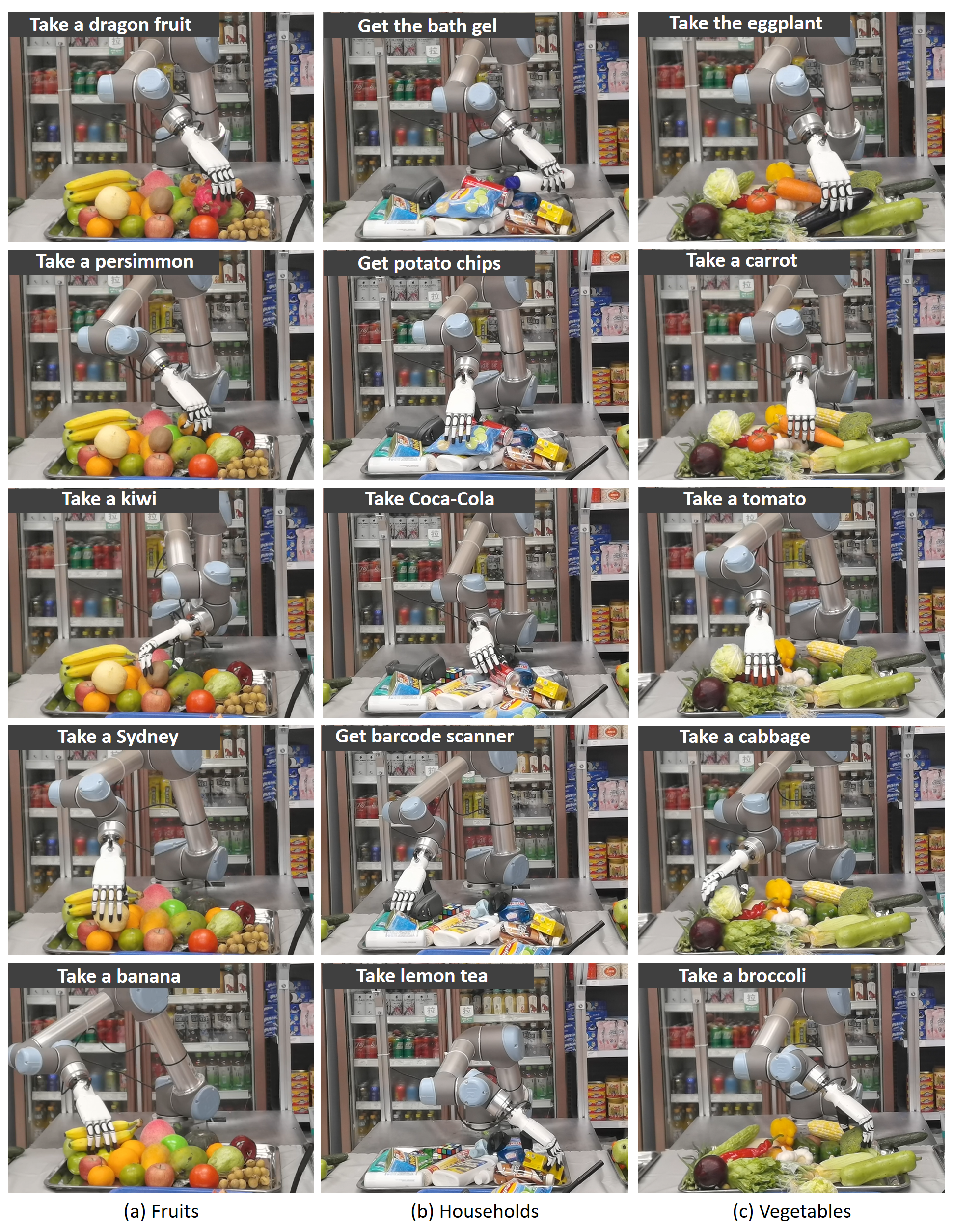}
      \caption{\textbf{Application Scenarios of Voice-Commanded Dexterous Grasping.} Experimental setups illustrating the system's performance in cluttered environments: (a) Fruits; (b) Household Items; (c) Vegetables. These scenarios highlight the challenges and versatility of the EDGS across different object categories in dynamic settings.}
      \label{fig:application_scenarios}
    \end{figure*}
    
    In our experiments, we evaluated the performance of the EDGS in realistic, cluttered environments, using objects from three distinct categories: fruits,households and vegetables. These objects were placed in dense, disorganized settings to replicate complex real-world conditions, as shown in Figure \ref{fig:application_scenarios}. The EDGS effectively processed voice commands, demonstrating its ability to interpret human intent and integrate both auditory and visual cues to make informed grasping decisions. Notably, the system maintained high success rates across a range of challenging grasping tasks, indicating its robustness and adaptability to diverse object types and cluttered scenarios, as shown in Table \ref{tab:application_comparison}. The results indicate that the EDGS maintained a high success rate across all object categories, with the highest performance observed in the fruits category 98.3\%. The vegetables category followed closely with a success rate of 94.2\%, while the everyday items category recorded a success rate of 95.8\%. This demonstrates that our system can function as a versatile and accessible robotic assistant, with significant potential for automation in industries requiring dexterous manipulation and intelligent interaction.
    
    To further understand the system's performance, we conducted an error analysis on the failed grasp attempts. The primary causes of failure were identified as follows: 1) Object Shape Complexity: Grasping attempts on irregularly shaped objects resulted in a higher failure rate, particularly in the everyday items category; 2) Environmental Clutter: High-density arrangements of objects led to difficulties in accurately identifying and segmenting target objects, contributing to grasping failures; 3) Dynamic Interaction: The system occasionally struggled with objects that required delicate handling, resulting in slips during the transfer phase.

    \begin{table}[ht]
    \centering
    \resizebox{0.9\columnwidth}{!}{%
    \begin{tabular}{l c c}
    \toprule
    \multicolumn{1}{c|}{Object Category} & {Total Attempts}  & {Success Rate} \\
    \midrule
    \multicolumn{1}{c|}{Fruits} & 120 & 98.3\% \\
    \multicolumn{1}{c|}{Households}  & 120 & 94.2\% \\
     \multicolumn{1}{c|}{Vegetables} & 120 & 95.8\% \\
     \multicolumn{1}{c|}{\textbf{Overall}} & \textbf{360} & \textbf{96.1\%} \\
    \bottomrule
    \end{tabular}%
    }
    \caption{\textbf{EDGS Performance in Application Scenarios.}}
    \label{tab:application_comparison}
    \end{table}
    
    \section{Limitation and Discussion}
    Despite significant progress, a considerable gap persists between current system and human-level performance in complex environments:
    First, while our system leverages RGB-D modalities and enriched multimodal information from RERE to achieve effective semantic-object alignment, it remains constrained by the absence of human-level multimodal sensory integration, which leads to significant challenges in accurately identifying and segmenting densely cluttered and overlapping objects.  
    Second, current segmentation method only delineates object boundaries, failing to establish structural-to-functional semantic mappings in object sense, thus hindering its capability in nuanced manipulation tasks. 
    Third, our hand's lack of fine-grained haptic sensing components impedes the system's dynamic adaptability to highly complex and altering scenarios. Besides, its relative deficiency in degrees of freedom also limits the solution space of higher-quality grasp poses, and results in its inabilities of delicate in-hand re-orientation.
    Fourth, this single-hand, single-arm system fails to achieve force closure for objects in larger size, which typically requires of coordinated or combined action of dual-hands, dual-arms system.
    
    Despite all these limitations, this system, as a cohesively integrated hand-eye-brain embodied humanoid system, enables modular deployment on various robotic platform and caters the demands of diverse application scenarios.   
    \section{Conclusion}
    In this paper, we introduced the EDGS, a novel framework that enables precise object grasping in complex environments through natural language commands. By employing the RERE method, EDGS enhances object segmentation accuracy and reduces ambiguities in human instructions using vision-language models. Our system integrates modules for generating and refining dexterous grasp candidates, ensuring robust and adaptable grasping strategies. Extensive experiments demonstrate that EDGS achieves high success rates and stability in real-world scenarios, effectively handling diverse grasping tasks. This work highlights the potential of voice-driven robotic systems to transform human-robot interactions, making intelligent robots more accessible for everyday applications. Future research will focus on optimizing grasp generation efficiency and enhancing the system's adaptability to a wider range of object shapes and materials, further bridging the gap between robotic capabilities and human dexterity.

    \section*{Acknowledgments}
    This work is partially supported by the Guangdong Basic and Applied Basic Research Foundation under Grant 2023A1515011347; in part by the Shenzhen Science and Technology Program under Grant JCYJ20220818103006012, Grant ZDSYS20220606100601002, and Grant KJZD20230923114810022; and in part by the Shenzhen Institute of Artificial Intelligence and Robotics for Society.

    {\small
    \bibliographystyle{ieee_fullname}
    \bibliography{egbib}
    }
    
    \end{document}